\documentclass[10pt,twocolumn,letterpaper]{article}

\usepackage[pagenumbers]{wacv} % To force page numbers, e.g. for an arXiv version

\usepackage{multirow}
\usepackage{graphicx}
\usepackage{amsmath}
\usepackage{amssymb}
\usepackage{booktabs}
\usepackage{lipsum}
\usepackage{makecell}
\usepackage{pifont}
\usepackage{float}

\usepackage[pagebackref,breaklinks,colorlinks]{hyperref}

\usepackage[capitalize]{cleveref}
\crefname{section}{Sec.}{Secs.}
\Crefname{section}{Section}{Sections}
\Crefname{table}{Table}{Tables}
\crefname{table}{Tab.}{Tabs.}

\def\wacvPaperID{11} % *** Enter the WACV Paper ID here
\def\confName{WACV}
\def\confYear{2025}

\begin{document}

\title{YOLO11-JDE: Fast and Accurate Multi-Object Tracking with Self-Supervised Re-ID}

\author{Iñaki Erregue$^{1}$\\
{\tt\small ierregal31@alumnes.ub.edu}
\and
Kamal Nasrollahi$^{3,4}$\\
{\tt\small kn@create.aau.dk}
\and
Sergio Escalera$^{1,2,3}$\\
{\tt\small sescalera@ub.edu}
\and
{\normalsize$^1$Universitat de Barcelona}
\and
{\normalsize$^2$Computer Vision Center}
\and
{\normalsize$^3$Aalborg University}
\and
{\normalsize$^4$Milestone Systems}
}
\maketitle

%%%%%%%%% ABSTRACT
\begin{abstract}
We introduce YOLO11-JDE, a fast and accurate multi-object tracking (MOT) solution that combines real-time object detection with self-supervised Re-Identification (Re-ID). By incorporating a dedicated Re-ID branch into YOLO11s, our model performs Joint Detection and Embedding (JDE), generating appearance features for each detection. The Re-ID branch is trained in a fully self-supervised setting while simultaneously training for detection, eliminating the need for costly identity-labeled datasets. The triplet loss, with hard positive and semi-hard negative mining strategies, is used for learning discriminative embeddings. Data association is enhanced with a custom tracking implementation that successfully integrates motion, appearance, and location cues. YOLO11-JDE achieves competitive results on MOT17 and MOT20 benchmarks, surpassing existing JDE methods in terms of FPS and using up to ten times fewer parameters. Thus, making our method a highly attractive solution for real-world applications. The code is publicly available at \url{https://github.com/inakierregueab/YOLO11-JDE}.
\end{abstract}

%%%%%%%%% BODY TEXT
\section{Introduction}
\label{sec:intro}
Multi-Object Tracking (MOT) is a fundamental task in computer vision that involves detecting multiple objects in a video sequence and maintaining their identities discriminated across frames. From autonomous driving \cite{RetinaTrack, DEFT, 3dcars} and video surveillance \cite{surveillance1,surveillance2} to sports analytics \cite{sports1,sports2,sports3} and robotics \cite{robotics2, robots1}, MOT is a key component for numerous real-world applications. Despite significant advancements in the field, factors such as frequent occlusions, complex and unpredictable motion patterns, and the need for real-time performance in practical scenarios remain challenging \cite{bashar2022multipleobjecttrackingrecent, Ciaparrone_2020}.

Among the different paradigms for MOT, the Tracking-by-Detection (TbD) approach has become the most widely used due to its modularity and flexibility, separating the task into two stages: detecting objects in each frame and associating these detections across consecutive frames to maintain identities. Many methods integrate Re-Identification (Re-ID) embeddings to ease the matching process. These appearance cues are particularly valuable in challenging scenarios involving occlusions or objects with similar motion patterns, as they provide an additional layer of discrimination beyond spatial and temporal information.
%To improve association, many methods integrate Re-Identification (Re-ID) embeddings, which provide appearance cues to better distinguish identities. These embeddings are particularly valuable in challenging scenarios involving occlusions or objects with similar motion patterns, as they provide an additional layer of discrimination beyond spatial and temporal information. 
\begin{figure}[t]
  \centering
   \includegraphics[width=\linewidth]{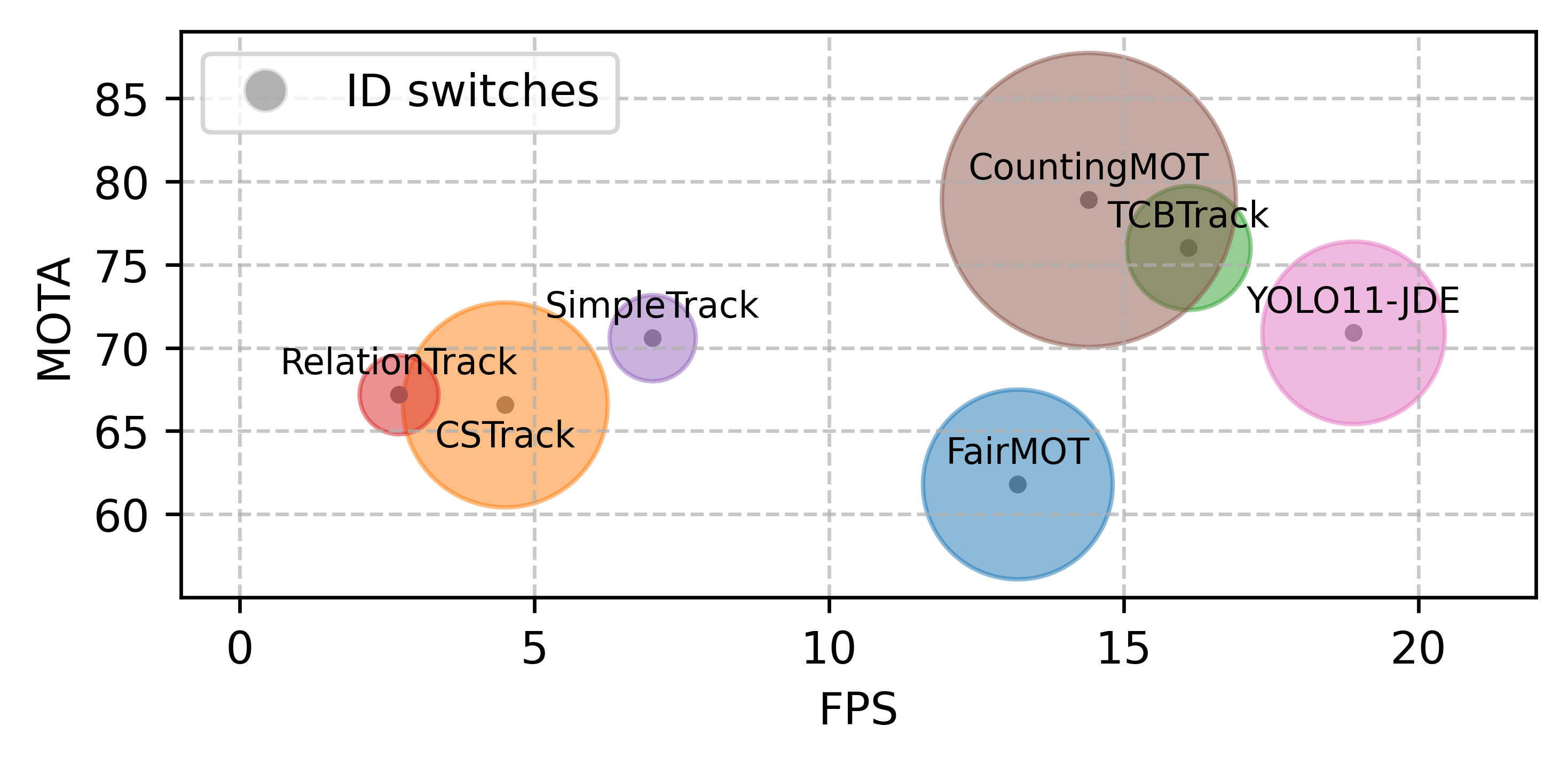}
   \vspace*{-7mm}
   \caption{Comparative analysis of JDE models on the MOT20 test set. MOTA and FPS on the vertical and horizontal axis respectively. ID Switches are represented by the bubble size. YOLO11-JDE achieves a strong balance between tracking performance and inference speed.}
   \label{fig:scatter}
\end{figure}

While substantial progress has been made in both detection and Re-ID fields, most methods adopt a two-stage approach, known as Separate Detection and Embedding (SDE), where detection and Re-ID are performed independently \cite{BoTSORT, SMILETrack, ConfTrack, BoostTrack}. While effective, these methods suffer from scalability issues due to the lack of feature sharing and the computational cost of applying the Re-ID model to every bounding box. To address these limitations, recent advancements introduced Joint Detection and Embedding (JDE) models, which unify object detection and Re-ID feature extraction processes into a single model \cite{originalJDE, FairMOT, CSTrack, OMCTrack, TCBTrack, RelationTrack, SimpleTrack, QDTrack, countingmot}. By sharing features between the two tasks and jointly optimizing them, JDE models significantly reduce computational overhead, making it an attractive paradigm for MOT. 

\begin{figure*}[t]
  \centering
  \captionsetup[subfigure]{labelformat=empty}%
  \subcaptionbox{\label{sub:1}}{}%
  \subcaptionbox{\label{sub:2}}{}%
  \subcaptionbox{\label{sub:3}}{}%
  \subcaptionbox{\label{sub:4}}{}%
   \includegraphics[width=0.9\linewidth]{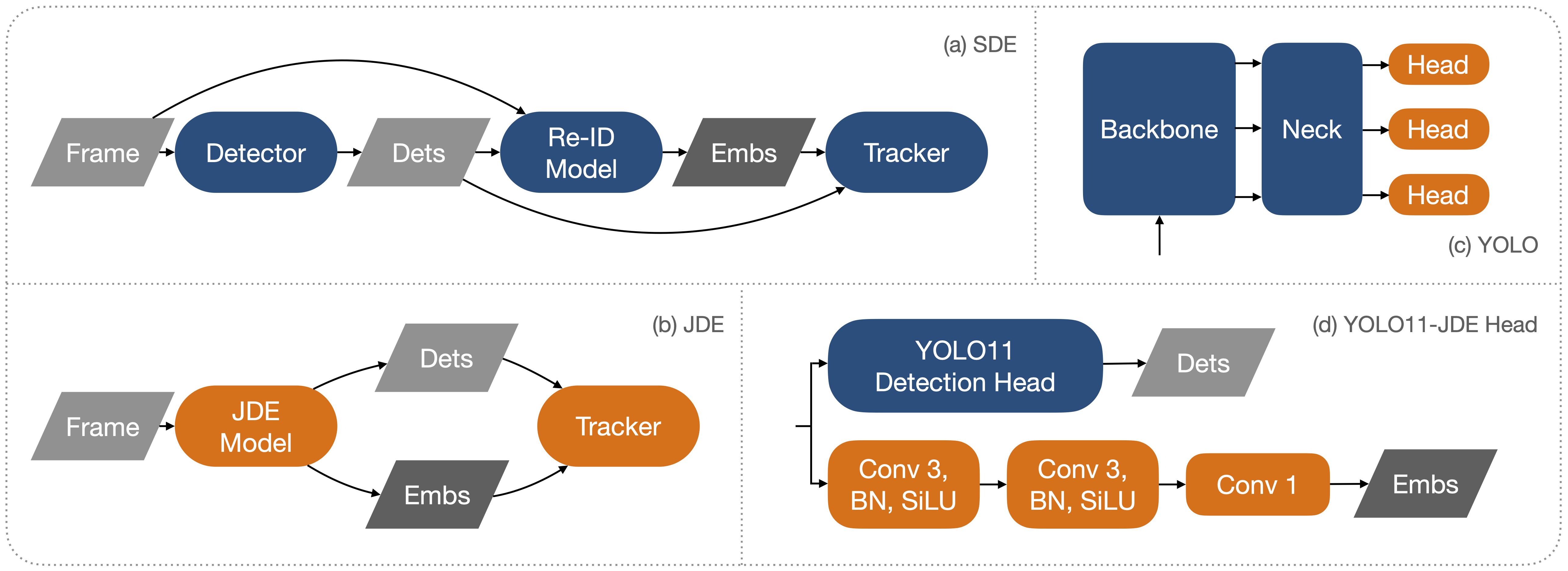}
   \vspace*{-5mm}
   \caption{Comparison of tracking architectures: (a) Separate Detection and Embedding (SDE), where detection and Re-ID are performed by separate models; (b) Joint Detection and Embedding (JDE), integrating detection and Re-ID into a single model; (c) Basic YOLO model structure including the backbone, neck, and multiple output heads; and (d) YOLO11-JDE head, featuring a specialized Re-ID branch. Grey parallelograms represent data, while rounded rectangles depict models. Our main contributions are highlighted in orange.}
   \label{fig:diagram}
\end{figure*}

Joint training of the detection and embedding tasks of JDE models pose unique challenges. While object detection focuses on clustering features to separate object classes, Re-ID requires some intra-class variability to achieve good discrimination between individual objects within the same class. This inherent conflict complicates the optimization process, making the choice of loss functions particularly critical in JDE models. Furthermore, achieving state-of-the-art performance often requires large-scale, labelled tracking datasets for supervision, which are expensive and time-consuming to obtain.

In this work, we present an end-to-end framework that builds upon the popular Ultralytics \cite{ultralytics} framework and the state-of-the-art detector, YOLO11 \cite{yolo11_ultralytics}, modified to perform joint detection and embedding. To tackle the inherent difficulties in joint training, we explore the field of deep metric learning, aiming to strike an optimal balance between detection and embedding objectives by using the well-established triplet loss \cite{tripletloss, tripletloss_origins}. Moreover, to mitigate the need for extensive identity label supervision, we utilize strong data augmentation techniques, particularly Mosaic data augmentation \cite{mosaic}, enabling our model to perform effectively in a fully self-supervised setting. Our approach drastically reduces the number of parameters compared to existing JDE methods, resulting in a notable speed-up in frames per second (FPS). Evaluated on the MOT Challenge benchmarks, YOLO11-JDE demonstrates competitive tracking accuracy while maintaining high efficiency (see \cref{fig:scatter}), making it well-suited for real-time MOT applications, where inference speed and model size are crucial. In summary, our main contributions are:
\begin{itemize}
    \item YOLO11-JDE, a modified YOLO11s that performs JDE, being small, fast and accurate.
    \item A self/semi-supervised setting for training JDE models based on Mosaic data augmentation and the triplet loss function.
    \item A customized data association algorithm that integrates motion, location and appearance cues.
\end{itemize}
\section{Related Work}
\label{sec:related}
\subsection{Tracking-by-Detection}
The task of MOT can be broadly categorized into three main paradigms based on how detection and tracking tasks are combined: tracking-by-regression, tracking-by-detection and tracking-by-attention. Nonetheless, TbD stands out as the most practical and widely used approach in both research and real-world applications. These trackers divide MOT into two separate tasks: detection and association. The tracking process begins with the identification of potential objects of interest in each frame using high-performance detectors like YOLOX~\cite{yolox}, Faster R-CNN~\cite{fasterrcnn}, or CenterNet~\cite{centernet}. Detected objects are then associated across consecutive frames using tracker algorithms that perform data association employing several cues (motion, location, appearance, etc.). 

Since the candidate boxes can be directly provided by off-the-shelf detectors, TbD methods mainly focus on improving the association performance. Early methods like SORT~\cite{SORT} employ a Kalman filter~\cite{KalmanFilter} to predict object positions in subsequent frames, assuming linear motion dynamics. Data association is performed using the Hungarian algorithm~\cite{Hungarian}, with a cost matrix based on the Intersection-over-Union (IoU) between predicted and detected bounding boxes. More recent advancements, like ByteTrack \cite{ByteTrack}, utilize all outputted detections, including low-confidence ones, in a two-stage cascade matching strategy. ConfTrack \cite{ConfTrack} and BoostTrack \cite{BoostTrack} go one step further by introducing novel penalization and boosting methods for low- and high-confidence detections in the matching process respectively. In a different direction, C-BIoU tracker \cite{CBIoUTracker} mitigates the effect of irregular motions by adding buffers to expand the matching space of detections and tracks. 

\subsection{Re-Identification}
To better deal with occlusions, crowded scenes, and non-linear motion, appearance similarity is commonly used in addition to IoU and motion cues. Thus, modern systems, like DeepSORT \cite{DeepSORT}, BoT-SORT \cite{BoTSORT}, SMILETrack~\cite{SMILETrack} and many others \cite{POI, ConfTrack, BoostTrack}, incorporate the extraction of discriminative Re-ID features for detected objects. These embeddings can be obtained either using an external high-quality feature extractor (\eg FastReID \cite{FastReID}), or using JDE models (see \cref{sub:1,sub:2}). Despite achieving superior performance, SDE approaches suffer from massive computation costs since the feature extractor network needs to perform forward inference on the image or feature map crop of each bounding box, thus limiting real-time applications.

\subsection{Joint Detection and Embedding}
JDE models perform object detection and Re-ID feature extraction in a single network in order to reduce inference time. Focusing on one-shot detectors, Wang \etal \cite{originalJDE} redesign the coupled prediction head of YOLOv3 \cite{yolov3} to extract embeddings of dimension 512 directly applying a 1 × 1 convolution layer on the shared features. Thus, ignoring the inherent differences between the three tasks involved. Moreover, \cite{originalJDE} trains the Re-ID task using a classification approach, where extracted embeddings are fed into a shared fully-connected layer to output the class-wise logits, and then cross-entropy loss is applied. In this method, annotations without identity labels are ignored. CSTrack \cite{CSTrack} adopts YOLOv5 \cite{yolov5} as detector and introduces two new modules to decouple the Re-ID task and fuse embeddings across scales. Subsequent advancements, such as OMC \cite{OMCTrack} and TCBTrack \cite{TCBTrack}, emphasize the temporal refinement of appearance cues.

On the other hand, FairMOT \cite{FairMOT} uses a modified version of the anchor-free detector CenterNet to output 128-dimensional features alongside each detection. Similarly to previously mentioned approaches, FairMOT learns Re-ID features through a classification task. In addition to the standard training strategy, FairMOT introduces a single-image training approach tailored for image-level object detection datasets. Each bounding box is assigned a unique identity, effectively treating every object instance in the dataset as a distinct class. By applying various transformations to the entire image, the model is exposed to each identity across multiple conditions. Despite reporting acceptable results, this self-supervised approach is used as a pre-training routine and not explored any deeper. QDTrack \cite{QDTrack} further investigates the self-supervised paradigm incorporating MixUp \cite{mixup} and Mosaic transformations, along with an extension of the InfoNCE loss \cite{InfoNCE} paired with a regularization term. Meanwhile, other models based on CenterNet, like RelationTrack \cite{RelationTrack} and SimpleTrack \cite{SimpleTrack} focus on decoupling both tasks and improving data association.

More recent JDE methods, CountingMOT \cite{countingmot} and UTM~\cite{utm}, achieve state-of-the-art performance on MOTChallenge benchmarks. The former, build upon FairMOT, adds an extra counting task to be shared across detection and density estimation branches, boosting the performance in crowded scenes. The latter, includes the data association step into a unified tracker model, creating a positive feedback loop boosting detection and Re-ID altogether.

Despite being designed for autonomous driving scenarios, RetinaTrack\cite{RetinaTrack} is also noteworthy. Designed on top of RetinaNet \cite{RetinaNet}, it performs the task of JDE using the triplet loss and mining hard triplets.

\section{YOLO11-JDE}
In this section we detail the technical aspects of YOLO11-JDE including its modified architecture, the different strategies employed for effectively training the Re-ID branch in a self-supervised fashion, and the integration of Re-ID embeddings into the online data association process.

\subsection{Architecture}
Following related JDE approaches like~\cite{originalJDE,CSTrack,TCBTrack, utm, QDTrack}, our framework is based on the YOLO family of detectors, which typically consist of a backbone for generating feature maps, a neck that refines them by fusing shallow and deep representations, and three prediction heads (see \cref{sub:3}). Particularly, the state-of-the-art version YOLO11s has been chosen for its efficiency, accuracy and real-time performance. We have incorporated a Re-ID branch in the original multi-task decoupled head, taking inspiration from the design of the bounding box and segmentation regression branches. The Re-ID branch processes input feature maps through two consecutive 3x3 convolutional layers, each followed by batch normalization and the SiLU activation function. A third 1x1 convolutional layer maps the features into the corresponding embedding dimension with no batch normalization applied, following best practices suggested in \cite{FastReID}. This simple yet effective design allows the Re-ID branch to learn discriminative features without introducing unnecessary complexity and assessing the task in a consistent manner with the other object detection tasks: classification and bounding box regression. Thus, YOLO11-JDE outputs an appearance embedding for each detection alongside its predicted class and bounding box (see \cref{sub:4}). 

\subsection{Self-Supervised Training Strategy}
\begin{figure}[t]
  \centering
   \includegraphics[width=1.0\linewidth]{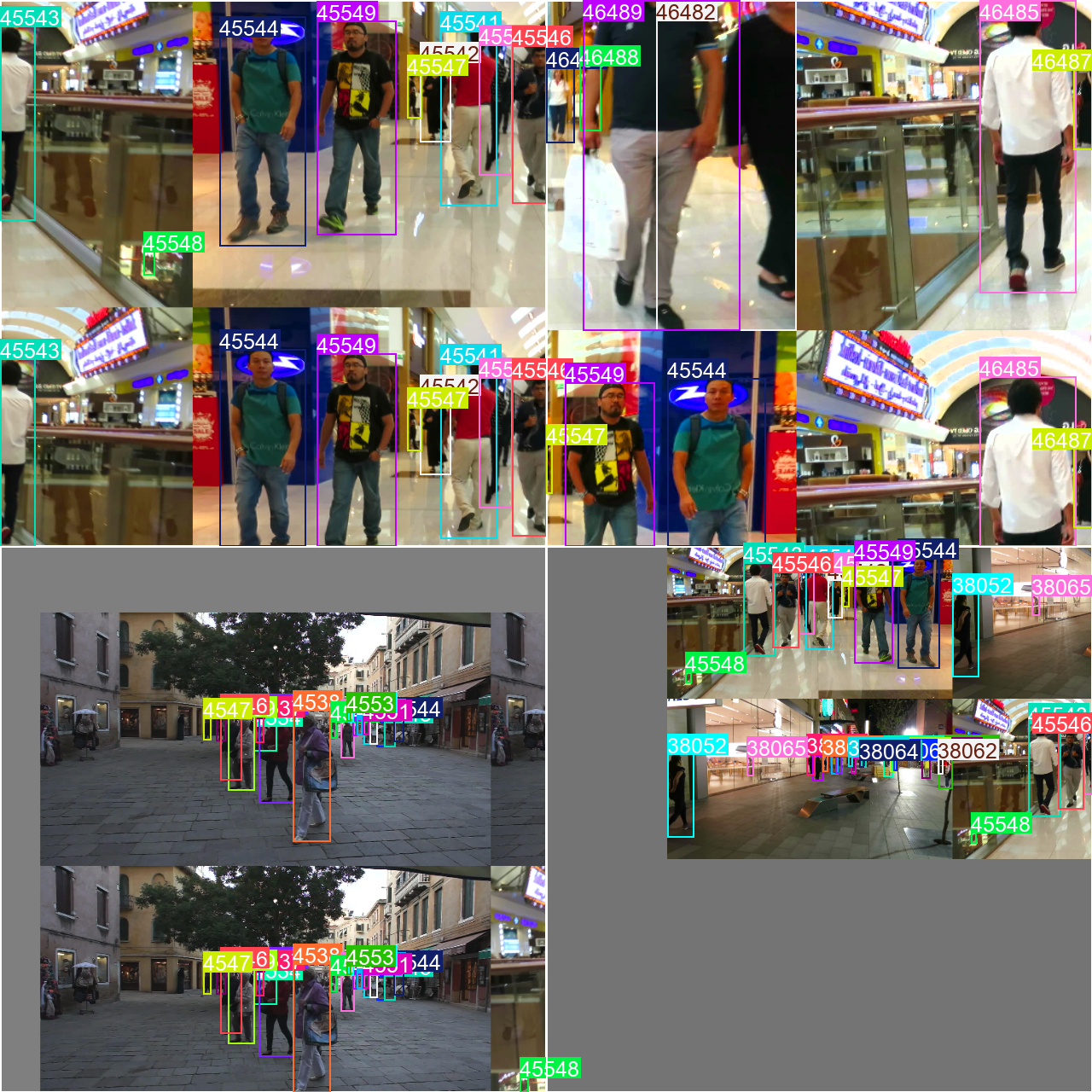}
   \caption{Example of four training images using Mosaic data augmentation for JDE. This technique combines multiple images, showing several identities (\eg, IDs 45543, 45544, 45549) under diverse transformations in a single input image and/or batch.}
   \label{fig:mosaic}
\end{figure}
The goal of the Re-ID branch is to produce robust embeddings that facilitate data association between consecutive frames, while minimizing the reliance on large-scale labelled tracking datasets. To achieve this, we aim for a fully self-supervised training approach, inspired by the work of FairMOT and QDTrack.

A core aspect of our self-supervised strategy is the use of Mosaic data augmentation \cite{mosaic}, a technique commonly employed in training modern object detectors like YOLO11. Mosaic augmentation works by combining four different image patches into a single input image, effectively enabling the model to review the same identities under diverse transformations, including variations in color, scale, rotation, etc. As depicted in \cref{fig:mosaic}, this approach allows the JDE model to learn robust features by exposing them to multiple augmented versions of the same identity within the same input image and/or batch while ordinarily training for detection. Thus, learning to output appearance features for each detection almost for free.

While our approach is intended to be fully self-supervised, it is also compatible with semi-supervised training, where a small amount of labelled tracking/identity data can complement the training procedure. This flexibility ensures the framework can adapt to scenarios with varying levels of data availability, which is crucial in real-world application.

\subsection{Re-ID Loss}
For a given training batch, the model outputs $N$ foreground predictions, each with an associated embedding that has a ground truth identity label assigned. The objective of the loss function is to pull embeddings with the same identity (positives) close together in the feature space while pushing those of different identities (negatives) further apart. This learning paradigm is a central concept in deep metric learning, where the goal is to learn a feature space where distances directly encode meaningful relationships between data points. The Re-ID task can be approached either as a classification problem or directly optimizing pair-wise relative distances between embeddings~\cite{dml_losses}.

Inspired by common Re-ID models trained in an contrastive fashion \cite{FastReID, MoCo, simclr, unsupervised_reid, unsupervised_reid2}, we adopt a pair-wise approach given its scalability to a large numbers of identities. While advanced pair-wise losses like Multi-Similarity \cite{multisim}, InfoNCE \cite{InfoNCE}, or Angular \cite{angular_loss} offer enhanced performance in certain tasks, we chose the triplet loss due to its simplicity, efficiency, and proven effectiveness \cite{defense_triplet}. The triplet loss aims to enforce a margin $m$ between positive and negative samples by ensuring that an anchor (a sample from a given identity) is closer to its positive counterpart than to a negative sample. The loss function is defined as:

\begin{equation}
  L_{triplet} = \sum_{\{a,p,n\}} [d_{ap} - d_{an} +m]_{+} \, ,
  \label{eq:triplet_loss}
\end{equation}
where $\{a,p,n\}$ represents the set of all triplets to be evaluated; and $d_{ap}$ and $d_{an}$ represent the distances between the anchor and the positive and negative samples respectively.%The notation $[x]_+$ ensures that the loss is only non-zero when the condition $d_{ap} +m < d_{an}$ is violated. 

Given $N$ embeddings, the total number of triplets scales as $O(N^3)$. This rapid growth makes it computationally unfeasible to use all possible combinations directly. Moreover, many of the resulting triplets would offer little new information to the model, slowing down convergence. To address these issues, recent advancements in pair-based metric learning have focused on more informative sampling strategies. In our setup, we use hard positive and semi-hard negative sampling strategy to obtain a total of $N$ triplets, although other strategies are also explored, \cref{subsec:ablations_loss}. On the one hand, hard positive mining chooses as positive for each anchor its furthest embedding with the same identity (most dissimilar). On the other hand, semi-hard negative mining selects the hardest negative sample per anchor (most similar embedding with different identity), such that it is further than the selected positive. Thus, selecting negatives pairs that are not too easy (far apart) but also not too difficult (close together). By utilizing these sampling strategies, we ensure that each triplet is informative and challenging, accelerating convergence, improving the overall performance and mitigating the issues of computational infeasibility.

\subsection{Data Association}
Initially, we adopted the two-stage online data association strategy used in FairMOT. Tracklets are initialized from detections in the first frame and updated in subsequent frames using a combination of motion and appearance cues. In the first stage, a Kalman Filter predicts tracklet locations and the Mahalanobis distance %$D_m$
is computed between predicted and detected boxes. Normalized Re-ID embeddings are use to compute a cosine distance matrix%$D_r$
, which is fused with the Mahalanobis distance to obtain the final cost matrix.
%\begin{equation}
%    D = \lambda D_r +(1-\lambda)D_m \, ,
%\end{equation}
%where $\lambda$ is set to 0.98. 
The matches are determined using the Hungarian algorithm. %with a matching threshold $\tau_1=0.4$. 
In the second stage, unmatched tracklets and detections are linked based on bounding box IoU with a stricter matching threshold%$\tau_2=0.5$
. Unmatched detections can initialize new tracks, while unmatched tracklets persist for 30 frames to handle occlusions. Following \cite{strongsort}, appearance features %$e_i$ 
are updated using an exponential moving average.
%\begin{equation}
%    e_i^t = \alpha e^{t-1}_i + (1-\alpha)f^t_i \, ,
%\end{equation}
%where $f^t_i$ is the appearance embedding of the current matched observation and the momentum term $\alpha$ is set to 0.9.

Building on FairMOT's tracker and inspired by ByteTrack, we have implemented a simple yet effective custom tracker for the YOLO11-JDE model. In the first stage, confident predictions are matched using a combination of motion, appearance, and localization cues. Motion is fused with appearance, following the approach of FairMOT, while also discarding matches with low IoU overlap. The IoU distance matrix is then combined with the confidence scores of detections, and low-similarity matches are discarded. The final cost matrix is a linear combination of these two factors. For low-confidence predictions and unmatched detections, linkage is performed using IoU alone. This approach balances computational simplicity with robust tracking performance.

\section{Experiments}
\subsection{Dataset and Metrics}
Seven datasets are commonly used when training JDE models on pedestrian tracking. Detection datasets include CrowdHuman \cite{CrowdHuman}, ETH \cite{ETH} and CityPersons \cite{CityPersons}, while MOT17 \cite{MOT17}, CalTech \cite{CalTech}, CUHK-SYSU \cite{cuhk} and PRW \cite{prw} also provide identity annotations. In our study we will only explore the mentioned object detection datasets with the exception of MOT17, which is added to fine-tune the model for the final evaluation. Following previous work \cite{BoostTrack, originalJDE}, we construct a MOT17 validation set using the second half of each training sequence and chop off the videos in ETH that are overlapped with the MOT16 \cite{mot16} benchmark.

We evaluate our approach on the testing sets of two widely recognized benchmarks, MOT17 and MOT20~\cite{MOT20}. For overall tracking accuracy, we primarily rely on HOTA~\cite{HOTA} due to its balanced evaluation of detection, association and trajectory quality. However, we also consider IDF1~\cite{IDF1} and MOTA from CLEAR metrics~\cite{MOTA} to provide additional insights into identity preservation and overall tracking performance. Detection performance is assessed using Average Precision (AP) with the common 50:95 acceptance IoU threshold range. While the quality of the Re-ID embeddings and the training convergence are monitored using clustering metrics like the Silhouette score~\cite{Silhouette}, retrieval mean Average Precision, and simpler indicators like the mean positive and negative Euclidean and cosine distances.

\subsection{Implementation Details}
Our framework builds on the Ultralytics infrastructure, modified to handle the task of JDE by incorporating identity labels management, a new JDE head, metrics for monitoring joint optimization, and a new set of tracking algorithms. Additionally, JDE loss functions and mining strategies are implemented using the PyTorch Metric Learning library \cite{PyTorchML}. Identity annotations are processed from existing datasets or generated synthetically if not available. They are preserved during data augmentation and foreground prediction alignment. All experiments use the YOLO11s model with COCO \cite{COCO} pre-trained weights. The default configuration of hyperparameters for optimization and data augmentation is used, except for Mosaic, which is applied throughout the whole training. 

\subsection{Ablative Studies}
In this section we present rigorous studies of four critical factors in YOLO11-JDE, including the Re-ID loss, the dimensionality of the appearance features, and the amount of training data and supervision needed. A simplified experimental setup has been employed to isolate and analyze the impact of these factors while maintaining computational feasibility. Specifically, we adopt the small variant of YOLO11 as the baseline model, trained for 30 epochs with a batch size of 32. The Re-ID branch employs the triplet loss with a unitary weight and outputs 128-dimensional embeddings. Training data is limited to CrowdHuman~\cite{CrowdHuman} and the detections from the training half of MOT17, all resized to 640 pixels. For validation, detection performance is evaluated on the validation splits of both datasets, while the Re-ID performance is exclusively assessed using the ground truth identity labels from MOT17. The tracker algorithm from FairMOT, with its default configuration, is used for evaluating the ablations, including an inference resolution of 1088×608 pixels. To ensure a comprehensive evaluation and to account for potential interactions between factors, a sequential approach has been taken, where the best-performing configuration from one ablation is used as the baseline for the next. Evaluation metrics are given in percent. The best results of each ablation are shown in \textbf{bold}.

\subsubsection{Re-ID Loss}
\label{subsec:ablations_loss}
\textbf{Mining Strategy.} Ablation experiments begin by selecting the best mining strategy for the triplet loss, which has been used with the default margin  $m=0.05$. Various mining strategies have been explored, incorporating hard, semi-hard, and easy pairs for both positives and negatives. The results, summarized in \cref{tab:mining_strategy}, indicate that hard positives and semi-hard negatives yields the best overall performance both in terms of tracking accuracy and the quality of the Re-ID embeddings. This is likely due to the balanced challenge it presents to the model. Semi-hard negatives, which are not overly difficult to separate, are crucial for refining the decision boundary without introducing training instability. Meanwhile, hard positives force the model to learn robust discriminative features, enhancing intra-class consistency. Easier strategies, specially for negatives, rarely violate the margin condition, causing the model to focus mainly on the detection task, \ie, higher MOTA.
\begin{table}
  \centering
  {\small{
  \begin{tabular}{@{}lccc@{}}
    \toprule
    Mining Strategy (+,-)     & HOTA & MOTA & IDF1  \\
    \midrule
    (Hard, Hard)        & 51.66 & 55.23 & 59.41 \\
    (Hard, Semi-hard)   & \textbf{55.91} & 56.04 & \textbf{65.31} \\
    (Hard, Easy)        & 51.01 & 56.22 & 58.76 \\
    (Semi-hard, Hard)   & 53.93 & 55.11 & 62.34 \\
    (Semi-hard, Easy)   & 50.46 & \textbf{57.03} & 57.90\\
    (Easy, Hard)        & 53.76 & 54.21 & 61.50\\
    (Easy, Semi-Hard)   & 55.72 & 56.76 & 65.25 \\
    \bottomrule
  \end{tabular}
  }}
  \caption{Ablation results comparing different mining strategies for the triplet loss, showing the impact on tracking performance and Re-ID embedding quality.}
  \label{tab:mining_strategy}
\end{table}

\textbf{Loss Margin.} The next set of experiments focuses on the impact of the margin value $m$ in the triplet loss function. As shown in \cref{tab:loss_margin}, several values were tested around the baseline, with $m=0.075$ yielding to the best performance in terms of HOTA, MOTA and IDF1. Two additional experiments were conducted using this margin. First, swapping the distance computation (\ie, using the positive-negative distance instead of the anchor-negative if the latter violates the margin more) downgraded the performance, likely because it weakens the impact of the mining strategy. Second, smoothing the loss function by replacing the Hinge function with the Softplus function led to a noticeable increase in detection performance, although it slightly lags behind in HOTA.
\begin{table}
  \centering
  {\small{
  \begin{tabular}{@{}lccc@{}}
    \toprule
    Loss Margin ($m$) & HOTA & MOTA & IDF1  \\
    \midrule
    0.025 & 56.03 & 56.81 & 65.40\\
    0.05  & 55.91 & 56.04 & 65.31 \\
    0.075 & \textbf{56.37} & 56.38 & 66.45 \\
    0.1 & 55.62 & 55.88 & 65.42\\
    0.075 (smooth) & 56.21 & \textbf{57.61} & \textbf{66.50}\\
    0.075 (swap) & 55.42 & 56.07 & 64.76\\
    \bottomrule
  \end{tabular}
  }}
  \caption{Ablation results comparing different margin values for the triplet loss, as well as its smoothed and swapping distance counterparts.}
  \label{tab:loss_margin}
\end{table}

\textbf{Confidence Filtering.} Following the margin analysis, we investigate the impact of filtering the embeddings used for triplet mining, focusing on a confidence-based selection. The default approach mines among all available embeddings, ensuring maximum coverage but potentially including noisy or low-confidence samples. Therefore, we try limiting the embeddings to the top 75\% and 50\% most confident predictions per batch. The results, summarized in \cref{tab:confidence_mining}, illustrate how the model performs better when using all predictions. This could be attributed to the additional diversity provided by lower-confidence samples, which may expose the Re-ID branch to a wider range of challenging cases, ultimately leading to more robust feature learning. 
\begin{table}
  \centering
  {\small{
  \begin{tabular}{@{}lccc@{}}
    \toprule
    Confidence Filtering (\%)  & HOTA & MOTA & IDF1  \\
    \midrule
    100 & \textbf{56.37} & 56.38 & \textbf{66.45} \\
    75 & 55.88 & 56.52 & 65.56\\
    50 & 55.64 & \textbf{56.69} & 64.34\\
    \bottomrule
  \end{tabular}
  }}
  \caption{Ablation results comparing different confidence filtering thresholds during the mining process.}
  \label{tab:confidence_mining}
\end{table}

\textbf{Loss Weight.} The last set of ablations evaluates the impact of using three different weight values for the Re-ID loss. The goal is to understand how varying the contribution of the triplet loss in the overall multi-task objective function influences tracking performance. \Cref{tab:loss_weight} displays how the unitary weight outperforms the other configurations. Across several experiments, we have observed a general tendency: the lower the magnitude of the loss function, the better the results in detection. This suggests that a low but efficient signal in the Re-ID loss is crucial for ensuring that the detection task is not harmed during joint training.
\begin{table}
  \centering
  {\small{
  \begin{tabular}{@{}lccc@{}}
    \toprule
    Loss Weight  & HOTA & MOTA & IDF1 \\
    \midrule
    0.5 & 55.92 & \textbf{56.59} & 65.24\\
    1 & \textbf{56.37} & 56.38 & \textbf{66.45} \\
    1.5 & 55.27 & 56.59 & 64.58\\
    \bottomrule
  \end{tabular}
  }}
  \caption{Ablation results comparing different weight values for the triplet loss.}
  \label{tab:loss_weight}
\end{table}

\subsubsection{Feature Dimension}
In this subsection, we investigate the effect of varying the dimensionality of the embedding features on joint optimization and the final tracking performance. By experimenting with dimensions of 64, 128, and 256, we aim to identify the optimal size that provides robust identity embeddings while maintaining computational feasibility. As depicted in \cref{tab:feat_dim}, size 128 achieves the best balance. Dimension of 64 produces the highest MOTA, likely due to the lower signal loss benefiting detection during joint training. Conversely, increasing the dimension to 256 leads to slight declines across all metrics, probably caused by overfitting or redundant information in the higher-dimensional space.
\begin{table}
  \centering
  {\small{
  \begin{tabular}{@{}lccc@{}}
    \toprule
    Feat. Dim.  & HOTA & MOTA & IDF1 \\
    \midrule
    64 & 56.27 & \textbf{58.31} & 66.03\\
    128 & \textbf{56.37} & 56.38 & \textbf{66.45} \\
    256 & 55.44 & 56.26 & 65.09\\
    \bottomrule
  \end{tabular}
  }}
  \caption{Ablation results on the Re-ID feature dimensionality.}
  \label{tab:feat_dim}
\end{table}

\subsubsection{Training Datasets}
To assess the effect of the different types of supervision and data used in the training of our JDE model, we conducted another set of experiments. As shown in \cref{tab:datasets}, the model trained on CrowdHuman alone achieves a strong MOTA score, but its lower HOTA score reflects the necessity of fine-tuning with MOT17. Interestingly, incorporating identity supervision does not lead to improvements in HOTA or IDF1, suggesting that the model effectively learns more discriminative features using a fully self-supervised approach. While adding additional datasets such as ETH and CityPersons enhance detection performance, they do not improve tracking metrics, highlighting that the quality and relevance of fine-tuning data are more critical than data diversity.
\begin{table}
  \centering
  {\small{
  \begin{tabular}{@{}lccc@{}}
    \toprule
    Training Data  & HOTA & MOTA & IDF1  \\
    \midrule
    CH                                  & 52.39 & \textbf{58.53} & 65.12 \\
    CH, MOT17\textsuperscript{*}         & \textbf{56.37} & 56.38 & \textbf{66.45} \\
    CH, MOT17                            & 55.21 & 57.06 & 64.54\\
    CH, ETH, CP, MOT17\textsuperscript{*}  & 56.20 & 57.38 & 65.96\\
    CH, ETH, CP, MOT17                     & 55.54 & 57.83 & 65.14\\
    \bottomrule
  \end{tabular}
  }}
  \caption{Ablation results on the self/semi-supervised approaches. Superscript\textsuperscript{*} means that no identity annotations are used.}
  \label{tab:datasets}
\end{table}
 
\subsection{Data Association}
With the most promising configuration identified, we have trained the model for a 100 epochs using a batch size of 64 and an input image resolution of 1280 pixels. We then focus on fine-tuning the hyperparameters involved in the data association step. This section compares the results of using the default FairMOT tracker with its original parameters against a fine-tuned version for the YOLO11-JDE model, using the MOT17 training split. The default tracker, without adaptation, may struggle with mismatched confidence distributions and feature representations, leading to suboptimal data association and tracking accuracy. As shown in \cref{tab:tracker}, fine-tuning the tracker to align with YOLO11-JDE’s specific outputs significantly enhances its overall effectiveness. Moreover, the custom YOLO11-JDE tracker outperforms both FairMOT trackers across all metrics, ensuring more precise data association by integrating motion, appearance, and localization cues.
\begin{table}
  \centering
  {\small{
  \begin{tabular}{@{}lccc@{}}
    \toprule
    Tracker  & HOTA & MOTA & IDF1  \\
    \midrule
    FairMOT & 56.89 & 57.84 & 66.89 \\
    FairMOT (fine-tuned) & 57.25 & 57.10 & 67.16\\
    YOLO11-JDE &  \textbf{60.06}  & \textbf{58.25} & \textbf{71.84} \\
    \bottomrule
  \end{tabular}
  }}
  \caption{Results using different data association algorithms.}
  \label{tab:tracker}
\end{table}

\subsection{Results on MOTChallenge}
  We compare our method to existing literature, focusing on online JDE models targeting real-time performance. During inference, we use the new YOLO11-JDE tracker at an input resolution of 1280 pixels. Results on MOT17 and MOT20 test sets under the private detection protocol are shown in \cref{tab:sota}. Despite being the only fully self-supervised method in the comparison, YOLO11-JDE demonstrates competitive performance across benchmarks and significantly outpaces its counterparts in terms of FPS.
  %On MOT17, YOLO11-JDE achieves a HOTA score of 56.6, MOTA of 65.8, and IDF1 of 70.3, running at an impressive 35.9 FPS. On MOT20, it reaches a HOTA of 53.1, MOTA of 70.9, and IDF1 of 66.4, with 18.9 FPS.
  When it comes to identity switches (IDs), YOLO11-JDE outperforms many of its competitors, demonstrating the discriminative power of the produced embeddings. Therefore, we attribute its lower performance in overall tracking to the limitations of the model's detection capabilities, rather than its re-identification ability. Furthermore, YOLO11-JDE has less than 10M parameters, while top-performing methods like CountingMOT rely on computationally expensive detectors such as YOLOX-X (100M parameters) or CenterNet (22M parameters).
  
\begin{figure*}[t]
  \centering
   \includegraphics[width=\linewidth]{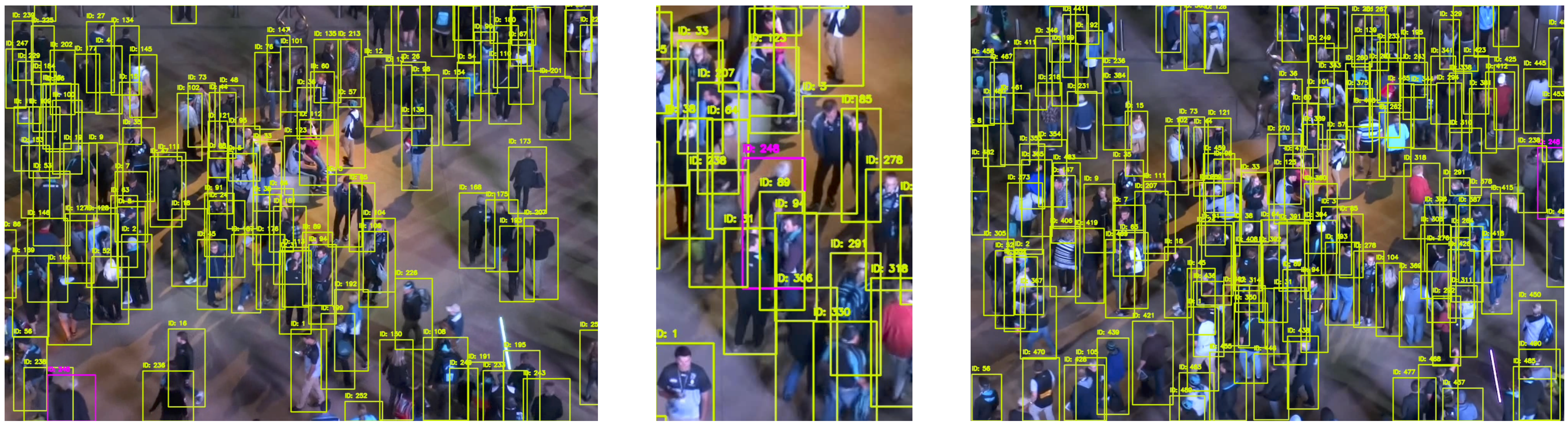}
   \caption{Example of consistent identity maintenance across frames on the MOT20-04 sequence despite multiple occlusions (ID 248).}
   \label{fig:occlusion}
\end{figure*}

  Interestingly, YOLO11-JDE performs better in MOT20 than in MOT17 when compared with its competitors. It is important to note that neither the YOLO11-JDE model nor the tracker have been trained using the MOT20 dataset. This improved performance on crowded scenes (see \cref{fig:occlusion}) can be attributed to the type of data used in training. The CrowdHuman dataset, which has a density of nearly 23 people per image, is magnified by Mosaic data augmentation, turning to approximately 90 people per image. This data composition makes YOLO11-JDE highly robust when handling crowded scenarios and partial occlusions.
\begin{table*}[t]
  \centering
  {\small{
  \begin{tabular}{@{}l c c c ccc cccc c@{}}
    \toprule
    \multirow{2}{*}{Tracker} & \multirow{2}{*}{Detector} & \multicolumn{5}{c}{MOT17} & \multicolumn{5}{c}{MOT20} \\
    \cmidrule(lr){3-7} \cmidrule(lr){8-12}
     & & HOTA & MOTA & IDF1 & IDs & FPS & HOTA & MOTA & IDF1 & IDs & FPS \\
    \midrule
    JDE \cite{originalJDE}            & YOLOv3         & -     & 63.0 & 59.5 & - & 18.8 & -    & -    & -    & -       \\
    FairMOT \cite{FairMOT}            & CenterNet      & 59.3  & 73.7 & 72.3 & 3303 & 25.9 & 54.6 & 61.8 & 67.3 & 5243 & 13.2    \\
    CSTrack \cite{CSTrack}            & YOLOv5          & 59.3  & 74.9 & 72.3 & 3567 & 15.8 & 54.0 & 66.6 & 68.8 & 3196 & 4.5     \\
    OMC \cite{OMCTrack}               & YOLOv5          & -     & 76.3 & 73.8 & - & 12.8  & -    & 70.7 & 67.8 & -& 6.7     \\
    TCBTrack \cite{TCBTrack}          & YOLOX-X         & 62.1  & \underline{79.3} & 75.8 & 2157 &\underline{27.7} & \underline{60.6} & \underline{76.0} & \underline{74.4} & \underline{1174} & \underline{16.1}    \\
    RelationTrack \cite{RelationTrack}& CenterNet       & 61.0  & 73.8 & 74.7 & \textbf{1374} & 7.4  & 56.5 & 67.2 & 70.5 & 4243 & 2.7     \\
    SimpleTrack \cite{SimpleTrack}    & CenterNet       & 61.0  & 74.1 & 75.7 & \underline{1500} & 22.5 & 56.6 & 70.6 & 69.6 & 2434 & 7.0     \\
    QDTrack \cite{QDTrack}            & YOLOX-X        & \underline{63.5}  & 78.7 & \underline{77.5} & 1935 & -  & 60.0 & 74.7 & 73.8 & \textbf{1042} & -       \\
    CountingMOT \cite{countingmot}    & YOLOX-X         & \textbf{63.6}  & \textbf{81.3} & \textbf{78.4} & 5118 & 26.4 & \textbf{63.6} & \textbf{78.9} & \textbf{78.6} & 1232 & 14.4 \\
    \midrule
    YOLO11s-JDE                 & YOLO11s       & 56.6 & 65.8  & 70.3  & 3177 & \textbf{35.9} & 53.1 & 70.9 & 66.4 &3091& \textbf{18.9} \\
    \bottomrule
  \end{tabular}
  }}
  \caption{Comparison with the state-of-the-art JDE models under the private detection protocol on MOT17 and MOT20 benchmarks. The best results are displayed in \textbf{bold} while the second best are \underline{underlined}.}
  \label{tab:sota}
\end{table*}
\section{Summary and Future Work}
In this work, we presented YOLO11-JDE, a lightweight and efficient MOT framework built upon YOLO11s, equipped with a Re-ID branch for joint detection and embedding. Our method demonstrates that Re-ID can be effectively trained in a fully self-supervised manner, avoiding the need for identity-labeled datasets while maintaining competitive tracking performance. By combining the triplet loss with hard positive and semi-hard negative mining strategies, YOLO11-JDE produces discriminative embeddings that are robust across various tracking scenarios, particularly crowded environments. Additionally, we developed a custom tracking algorithm that integrates motion, appearance, and location cues, effectively improving data association and aligning seamlessly with YOLO11-JDE's outputs. Evaluations on the MOT17 and MOT20 benchmarks highlight the method's ability to deliver comparable accuracy to state-of-the-art models while achieving superior FPS and using significantly fewer parameters. These qualities make YOLO11-JDE a practical and scalable solution for real-world applications.

For future work, we aim to address the limitations observed in detection performance by refining the architecture to better decouple Re-ID and detection tasks. Further improvements to appearance features, such as incorporating multi-scale embedding fusion, could enhance Re-ID robustness. Additionally, we plan to investigate the impact of stronger data augmentations, including rotations, shear and perspective transformations, Mixup and random patch erasing within bounding boxes.\\

\textbf{Acknowledgements.} This work has been partially supported by the Spanish project PID2022-136436NB-I00 and by ICREA under the ICREA Academia programme, and by the Milestone Research Program at the University of Barcelona.

%%%%%%%%% REFERENCES
{\small
\bibliographystyle{ieee_fullname}
\bibliography{egbib}
}

\end{document}